\documentclass{article}

    \PassOptionsToPackage{numbers, compress}{natbib}

\usepackage[preprint]{neurips_2022}
\usepackage[utf8]{inputenc} 
\usepackage[T1]{fontenc}    
\usepackage{hyperref}       
\usepackage{url}            
\usepackage{booktabs}       
\usepackage{amsfonts}       
\usepackage{nicefrac}       
\usepackage{microtype}      
\usepackage{xcolor}         
\usepackage{soul,color}
\usepackage{wrapfig}
\usepackage{caption}
\usepackage{subcaption}

\usepackage{graphicx}
\usepackage{amsmath}
\usepackage{amssymb}
\usepackage{mathtools}
\usepackage{amsthm}

\usepackage{booktabs}

\usepackage{amsmath,amsfonts,bm}

\def\secref#1{section~\ref{#1}}

\def\eqref#1{equation~\ref{#1}}

\def\1{\bm{1}}

\def\rvy{{\mathbf{y}}}

\def\rmH{{\mathbf{H}}}

\def\rmW{{\mathbf{W}}}

\def\rmZ{{\mathbf{Z}}}

\def\mA{{\bm{A}}}

\def\mH{{\bm{H}}}

\def\mM{{\bm{M}}}

\def\mU{{\bm{U}}}

\def\mW{{\bm{W}}}
\def\mX{{\bm{X}}}

\def\mZ{{\bm{Z}}}

\DeclareMathAlphabet{\mathsfit}{\encodingdefault}{\sfdefault}{m}{sl}
\SetMathAlphabet{\mathsfit}{bold}{\encodingdefault}{\sfdefault}{bx}{n}

\def\gE{{\mathcal{E}}}

\def\gG{{\mathcal{G}}}
\def\gH{{\mathcal{H}}}

\def\gW{{\mathcal{W}}}

\def\gZ{{\mathcal{Z}}}

\def\sT{{\mathbb{T}}}

\newcommand{\KL}{D_{\mathrm{KL}}}

\usepackage{url}
\usepackage{dsfont}
\usepackage{multirow}

\def\log{\mathrm{log}}

\usepackage{lipsum}
\usepackage{cuted}

\usepackage[capitalize,noabbrev]{cleveref}

\theoremstyle{plain}

\theoremstyle{definition}

\theoremstyle{remark}

\usepackage[textsize=tiny]{todonotes}

\title{Bayesian Graph Contrastive Learning}

\author{%
  Arman Hasanzadeh~\textsuperscript{\rm 1},  Mohammadreza Armandpour~\textsuperscript{\rm 1}, Ehsan Hajiramezanali \textsuperscript{\rm 1}\\ \textbf{Mingyuan Zhou}~\textsuperscript{\rm 2}, \textbf{Nick Duffield} \textsuperscript{\rm 1}, \textbf{Krishna Narayanan} \textsuperscript{\rm 1}\\
\textsuperscript{\rm 1}Texas A\&M University \qquad \textsuperscript{\rm 2}The University of Texas at Austin\\ 
{\tt\small \textsuperscript{\rm 1}\{armanihm, armandpur1990, ehsanr, duffieldng, krn\}@tamu.edu, \textsuperscript{\rm 2}mzhou@utexas.edu %
}
}

\begin{document}

\maketitle

\begin{abstract}
Contrastive learning has become a key component of self-supervised learning approaches for graph-structured data. Despite their success, existing graph contrastive learning methods are incapable of uncertainty quantification for node representations or their downstream tasks, limiting their application in high-stakes domains. In this paper, we propose a novel Bayesian perspective of graph contrastive learning methods showing random augmentations leads to stochastic encoders. As a result, our proposed method represents each node by a distribution in the latent space in contrast to existing techniques which embed each node to a deterministic vector. By learning distributional representations, we provide uncertainty estimates in downstream graph analytic tasks and increase the expressive power of the predictive model. In addition, we propose a Bayesian framework to infer the probability of perturbations in each view of the contrastive model, eliminating the need for a computationally expensive search for hyperparameter tuning. We demonstrate the performance of our model on uncertainty quantification, interpretability of latent representation and downstream predictive performance, and compare with related models in the literature.
\end{abstract}

\section{Introduction}\label{sec:intro}
Self-supervised contrastive methods have shown great promise to produce high-quality node representations by learning to be invariant to different augmentations in graph domain \cite{sun2019infograph,zhu2020grace,you2020graph,you2021graph}.
While these contrastive learning methods have achieved great empirical performance, to the best of our knowledge, they are limited to learn \emph{deterministic} node embeddings.
Such deterministic representations lack the capability of modeling uncertainty,
which is a natural consideration when having multiple information sources, i.e. node attributes and graph structure, and is of crucial importance in many real-life applications such as healthcare and autonomous driving \cite{kendall2017uncertainties,mukhoti2018evaluating}.
Another main challenge in Graph Contrastive Learning (GCL) methods is the proper choice of augmentation types and their hyperparameters. 
There has been recent works that attempt to address this issue by either assuming augmentation operation to be an invertible function \cite{tian2020makes} or only learning the probability of picking a specific type of augmentation and do not learn the parameters of the augmentations \cite{you2021graph}.

In this paper: 1) Instead of applying augmentations over input to make different views of data for GCL training, we propose a novel generalized augmentation which is applied to every layer of encoders. Our proposed method can be viewed as a stochastic regularization technique rather than an augmentation method. We theoretically prove that this is equivalent to having stochastic encoders. 
We further show that existing augmentation techniques are special cases of the proposed generalized augmentation, eliminating the need for choosing augmentation type.
2) We propose Bayesian Graph Contrastive Learning (BGCL) which is the first model in contrastive learning literature that 
enables Bayesian interpretability and provides uncertainty estimates for latent representations and downstream predictions. We theoretically prove that a GCL model with our generalized augmentation is an approximation of a Bayesian model, hence equipped with natural uncertainty quantification capabilities. 
3) We propose the first Bayesian approach for learning parameters of augmentation. Instead of a computationally expensive grid search for finding the optimal augmentation hyperparameters in existing works \cite{you2021graph}, BGCL learns the parameters of augmentation together with node embeddings (end-to-end training). 
We empirically demonstrate advantages of BGCL over existing methods on several benchmark dataset, in terms of uncertainty quantification and interpretability of latent representation.

\section{Graph contrastive learning}\label{sec: gcl}
Inspired by the success of contrastive methods in vision, graph contrastive learning methods have been introduced to learn node embeddings by minimizing a contrastive loss between two \emph{augmented views} of the same graph. \emph{Random corruptions} are applied to input graph to obtain augmented views. The corruptions are chosen form a set of corruption types which usually includes $\sT = \{\mathrm{NodeDrop}, \, \mathrm{EdgeDrop},\, \mathrm{FeatureDrop},$ $\mathrm{SubgraphSelection}\}$.
Although several graph contrastive losses have been proposed \cite{velickovic2018dgi,zhu2020grace,thakoor2021bootstrapped},  
the main goals of these objective functions are: 1) enforcing the representations of the same node in the two views of the graph to be close to each other, and 2) pushing apart representations of every other node.

To achieve the first goal, the vast majority of the methods include a term that measures the similarity between the representations of the same nodes in the two views. 
For the second goal, different approaches have been proposed. For example, GRACE \cite{zhu2020grace} deployed negative samples, like SimCLR \cite{chen2020simple}, while BGRL \cite{thakoor2021bootstrapped} uses different encoders for each view in the same way as BYOL \cite{grill2020byol}.
Without loss of generality, we carry out our analysis based on GRACE. We note that the same analysis could be applied to other methods as well. 

Given a graph $\gG = (\mA, \mX)$ with $\mA \in \mathbb{R}^{N \times N}$ as adjacency matrix and $\mX \in \mathbb{R}^{N \times F_0}$ as attributes matrix, GRACE generate two views of the graph $\tilde{\gG}_o = (\tilde{\mA}_o, \tilde{\mX}_o)$ and $\tilde{\gG}_t = (\tilde{\mA}_t, \tilde{\mX}_t)$ by sampling from a random corruption function. Subscripts $o$ and $t$ are used for indexing two views. Next, a graph neural networks (GNN) $f$, is deployed to map the augmented graphs to latent representations, $\mH_o$ and $\mH_t$, independently. The parameters of encoder (i.e. $f$) are learned by minimizing the following loss function:
\begin{equation}
\begin{split}
	&\mathcal{L}_{\text{GRACE}} = \underbrace{\frac{-1}{2N} \sum_{i=1}^{N}[\ell(\mH_{o,[i,:]}, \mH_{t,[i,:]}) + \ell(\mH_{t,[i,:]}, \mH_{o,[i,:]})]}_{\mathcal{L}_{\text{CNT}}} + \underbrace{\lambda \sum_{j=1}^{L} ||\mW^{(j)}||_2^2}_{\mathcal{L}_{\text{WD}}};\\
	&\qquad\qquad\ell(\mH_{o,[i,:]}, \mH_{t,[i,:]}) = \log \frac {e^{g\left(\mH_{o,[i,:]}, \mH_{t,[i,:]} \right) / \tau}} {e^{g\left(\mH_{o,[i,:]}, \mH_{t,[i,:]} \right) / \tau} + B_1 + B_2};\\
	& B_1 = \sum\limits_{k=1}^{N} \mathds{1}_{[k \neq i]} e^{g\left(\mH_{o,[i,:]}, \mH_{t,[k,:]} \right) / \tau},
	\qquad B_2 = \sum\limits_{k=1}^{N} \mathds{1}_{[k \neq i]} e^{g\left(\mH_{o,[i,:]}, \mH_{o,[k,:]} \right) / \tau},
	\label{equ: bgrl_loss}
\end{split}
\end{equation}
where $\tau$ is a hyperparameter, $\lambda$ is a $\ell^2$ weight decay hyperparameter, $g$ is a similarity function, $L$ is the number of layers in encoder, and $\mW^{(j)}$ represents the parameters of the encoder at $j$-th layer.  
It has been shown that $\mathcal{L}_{\text{CNT}}$ in the above equation is a lower bound for InfoNCE objective \cite{zhu2020grace}.

\section{Method}
\label{sec:method}
In this section, we introduce 
Bayesian Graph Contrastive Learning. 
Our proposed method: 1) infers an \emph{implicit} distribution as latent representation of each node, as opposed to deterministic vector representations in existing methods, allowing to capture uncertainty; 2) models augmentation masks as discrete random variables and infers their distribution using variational inference; 
3) provides an 
approach to translate the uncertainty of representation to the downstream predictive models. 

\subsection{Bayesian graph contrastive learning}
We are interested in developing a Bayesian interpretation of contrastive models which can be used to quantify uncertainty of representations. 
To that end, we will: 1) introduce a generalized augmentation model and show how the stochasticity of augmentations can be perceived as randomness in the parameters of the encoders;
2) show that training a graph contrastive model with our generalized augmentation model is an approximation of a Variational Graph Auto-Encoder (VGAE) \cite{kipf2016variational} with 
stochastic functions as encoders.

\begin{figure}[!t]
    \centering
    \includegraphics[width=0.93\columnwidth]{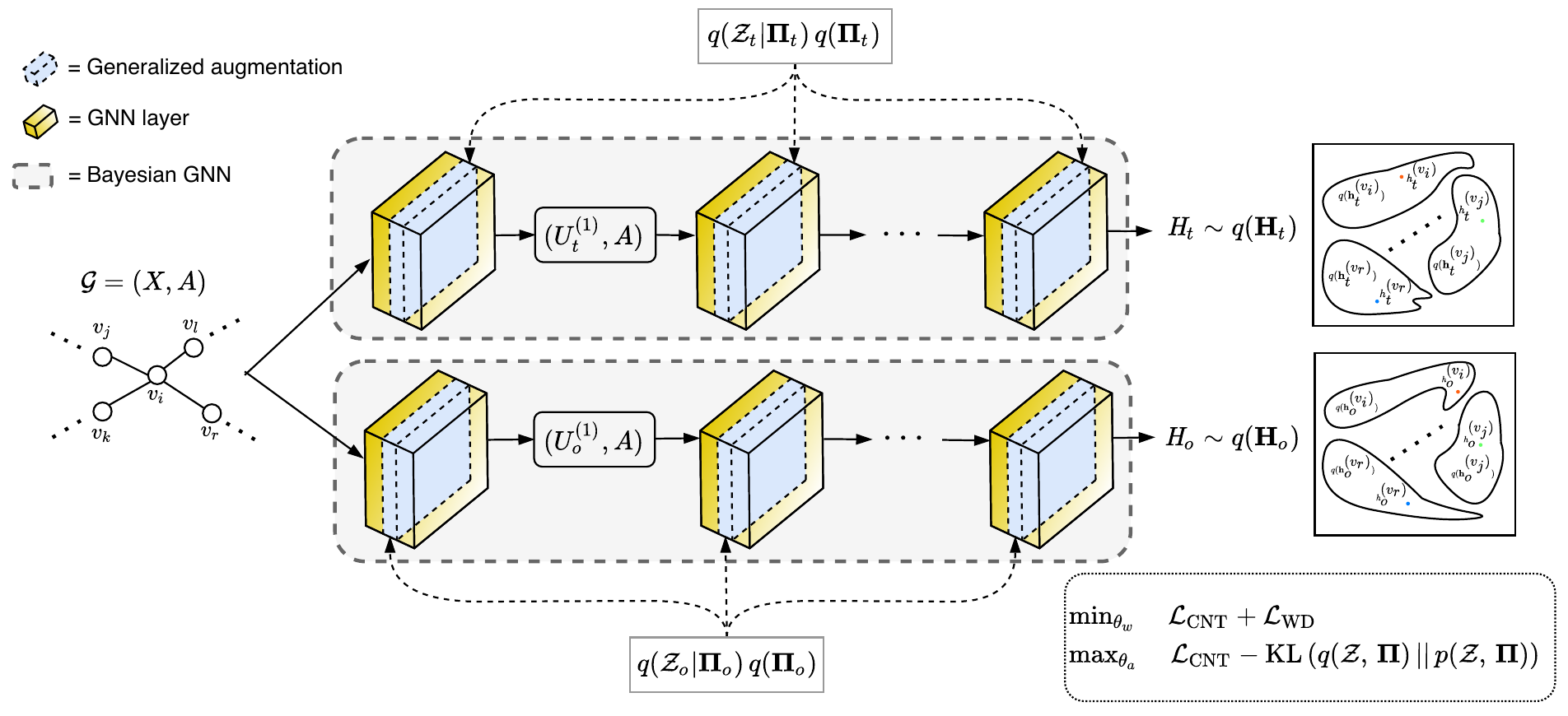}
    \caption{A schematic illustration of our proposed Bayesian graph contrastive learning. 
    }%
    \label{fig: schem}
\end{figure}

\subsubsection{Generalized augmentation model} \label{sec: gen_aug}
We first define a generalized augmentation method that includes the existing augmentation types as special cases. We note that all of the augmentations in $\sT$ (\secref{sec: gcl}) are simply parameterized by a random binary matrix. For example, $\mathrm{FeatureDrop}$ can be formulated as $\tilde{\mX} = \mX \odot \tilde{\mZ}$, where $\odot$ is the Hadamard product, and $\tilde{\mZ}$ is a sample drawn from a random Bernoulli matrix.

Unlike existing works where augmentations are only applied to the first layer (i.e. input data) of encoders, we extend the notion of augmentation and propose to multiply the output of each layer of encoders by a random mask. These masks could be drawn from different types and/or have different probability distributions.
This extension allows us to view augmentation as a stochastic regularization technique for GNNs. Indeed, conventional augmentation is a stochastic regularization that is only
applied to the first layer of encoders.

Furthermore, to increase flexibility of the model, we propose to multiply a different random mask with the adjacency matrix for every input-output feature pair \cite{hasanzadeh2020gdc}. 
More specifically, the $(l+1)$-th layer of graph neural network in an encoder with generalized augmentation is defined as follows:
\begin{equation}
    \mU^{(l+1)}_{[:,j]} = \sigma \left(\sum_{i=1}^{F_l} g_{a} \left(\mA \odot \tilde{\mZ}^{(l, i, j)} \right)\, \mU^{(l)}_{[:,i]} \, W^{(l)}_{[i,j]}  \right); \qquad \text{for}\quad j = 1,\, \dots \,, F_{l+1},\nonumber
    \label{equ: gdc_matrix}
\end{equation}
where $\mU^{(l)} \in \mathbb{R}^{N \times F_l}$ is the matrix of features at layer $l$ with $\mU^{(0)} = \mX$, $F_l$ is the number of features at layer $l$, $\sigma$ is the activation function, $g_a$ is the adjacency matrix normalization operator \cite{kipf2016semi}, and the elements of $\tilde{\mZ}^{(l, i, j)}$ are drawn from $\mathrm{Bernoulli}(\pi^{(l)})$. It can be easily shown that existing augmentations are special cases of our proposed method. For example, choosing $\tilde{\mZ}^{(l,i,j)} = \mathds{1}_{N \times N} \, \mathrm{diag}(\tilde{\mZ}_{[:,i]}^{(l)})$ and $\pi^{(0)} \neq 1$ while $\{\pi^{(l)} = 1\}_{l=1}^{L-1}$ leads to $\mathrm{FeatureDrop}$. We emphesize that our proposed generalization eliminates the need for choosing augmentation type.

Having defined our generalized augmentation model, we show that the randomness could be transferred from input to parameters of encoder. To that end, we rewrite and reorganize \eqref{equ: gdc_matrix} to have a node-wise view of the layers. More specifically,
\begin{equation}
    U^{(l+1)}_{[v,j]} 
    = \sigma \left(c_v \sum_{i=1}^{F_l} \sum_{u \in \{\mathcal{N}(v), v\}} U^{(l)}_{[u,i]} \left( \tilde{Z}_{[v,u]}^{(l, i, j)} \, W^{(l)}_{[i,j]} \right) \right) = \sigma \left(c_v \sum_{i=1}^{F_l} \sum_{u \in \{\mathcal{N}(v), v\}} U^{(l)}_{[u,i]} \, \tilde{W}^{(l,u,v)}_{[i,j]} \right),
\label{equ: gdc_node}
\end{equation}
where $c_v$ is a constant depending on the graph structure, $\mathcal{N}(v)$ is the first hop neighborhood of $v$, and $\tilde{W}^{(l,u,v)}_{[i,j]} := \tilde{Z}_{[v,u]}^{(l, i, j)} \, W^{(l)}_{[i,j]}$. This can be interpreted as learning a different random weight for each 
connection in a GNN layer. 
We refer to $\tilde{\rmW}^{(l,u,v)} = [\tilde{W}^{(l,u,v)}_{[i,j]}]$ as connection-specific weights.

\subsubsection{Bayesian interpretation}\label{sec: elbo}
Bayesian models offer a mathematically grounded framework to reason about model uncertainty \cite{gal2016dropout}. If we show that GCL with generalized augmentation is an approximation of a Bayesian model, then we can quantify model uncertainty. To that end, we show that a GCL with generalized augmentation is a special case of VGAE by proving $\mathcal{L}_{\text{CNT}} + \mathcal{L}_{\text{WD}}$ (\eqref{equ: bgrl_loss}) is a valid ELBO.

Consider a VGAE with two latent variables $\rmH_o$ and $\rmH_t$ and two graph neural networks, with connection-specific weights $\gW = \gW_o \cup \gW_t$ where $\gW_o = \{\gW_o^{(l)}\}_{l=1}^{L} = \{\{\tilde{\rmW}_o^{(l,u,v)}\}_{(u,v) \in \gE}\}_{l=1}^{L}$ and $\gW_t = \{\gW_t^{(l)}\}_{l=1}^{L} = \{\{\tilde{\rmW}_t^{(l,u,v)}\}_{(u,v) \in \gE}\}_{l=1}^{L}$, as encoders. 
Furthermore, assume that the generative model is factorized as $p(\gG, \rmH_o, \rmH_t, \gW_o, \gW_t) = p(\gG \,|\, \rmH_o, \rmH_t, \gW)\, p(\rmH_o, \rmH_t \,|\, \gW)\, p(\gW_o)\, p(\gW_t)$, and the approximate posterior is factorized as $q(\rmH_o, \rmH_t, \gW_o, \gW_t \,|\, \gG) = q(\rmH_o \,|\, \gG, \gW_o)\, q(\rmH_t \,|\, \gG, \gW_t)\, q(\gW_o)\, q(\gW_t)$. 
We can write the ELBO for this VGAE as follows:

\begin{equation}
\begin{split}
    \log\, p(\gG) =& \log\, \mathbb{E}_{q(\rmH_o, \rmH_t, \gW \,|\, \gG)} \left[\frac{p(\gG, \rmH_o, \rmH_t, \gW_o, \gW_t)}{q(\rmH_o, \rmH_t, \gW_o, \gW_t \,|\, \gG)} \right]\\
    \geq& \mathbb{E}_{q(\rmH_o, \rmH_t, \gW \,|\, \gG)} \log \Big[
    \,\frac{p(\gG \,|\, \rmH_o, \rmH_t, \gW)\, p(\rmH_o, \rmH_t \,|\, \gW)}{q(\rmH_o \,|\, \gG, \gW_o)\, q(\rmH_t \,|\, \gG, \gW_t) } \Big]
    + \mathbb{E}_{q(\gW)}\left[\log \frac{p(\gW_o)\, p(\gW_t)}{q(\gW_o)\, q(\gW_t)} \right]\\
    =& \mathbb{E}_{(\rmH_o, \rmH_t, \gW \,|\, \gG)} \log \left[\frac{p(\gG \,|\, \rmH_o, \rmH_t, \gW)\, p(\rmH_o, \rmH_t \,|\, \gW)}{q(\rmH_o \,|\, \gG, \gW)\, q(\rmH_t \,|\, \gG, \gW) } \right]\\
    &\qquad - \KL\left(q(\gW_o) \,||\, p(\gW_o)\right) - \KL\left(q(\gW_t) \,||\, p(\gW_t)\right)\\
    =& \mathcal{L}_{\text{VGAE}}.
    \label{equ: vgae}
\end{split}
\end{equation}
Note that we used Jensen's inequality in the derivation of ELBO above. Following the same procedure as \citet{aitchison2021infoncevae}, we substitute the following likelihood:
\begin{equation}
\begin{split}
    &\qquad\qquad\qquad p(\gG \,|\, \rmH_o, \rmH_t, \gW) = \frac{q(\rmH_o \,|\, \gG, \gW_o)\, q(\rmH_t \,|\, \gG, \gW_t)\, p_{\text{true}}(\gG)}{q(\rmH_o)\, q(\rmH_t)};\\
    &q(\rmH_o) = \mathbb{E}_{p_{\text{true}}(\gG)\, q(\gW_o)} q(\rmH_o \,|\, \gG, \gW_o),
    \quad q(\rmH_t) = \mathbb{E}_{p_{\text{true}}(\gG)\, q(\gW_t)} q(\rmH_t \,|\, \gG, \gW_t),
    \label{equ: ss-vgae-likelihood}
\end{split}
\end{equation}
in the definition of $\mathcal{L}_{\text{VGAE}}$ and derive:
\begin{equation}
\begin{split}
    \mathcal{L}_{\text{BGCL}}^{w} &= \mathbb{E}_{(\rmH_o, \rmH_t, \gW \,|\, \gG)} \log \left[\frac{p(\rmH_o, \rmH_t \,|\, \gW)}{q(\rmH_o)\, q(\rmH_t) } \right] \\
    & \qquad - \KL\left(q(\gW_o) \,||\, p(\gW_o)\right) - \KL\left(q(\gW_t) \,||\, p(\gW_t)\right) + \log\, p_{\text{true}}(\gG).
    \label{equ: ss-vgae}
\end{split}
\end{equation}
The last term in the above equation, i.e. $\log\, p_{\text{true}}(\gG)$, is independent of parameters, hence it does not affect the optimization and could be treated as a constant. 
Next, we parameterize the prior distribution latent representations in the same way as \citet{aitchison2021infoncevae}:
\begin{equation}
\begin{split}
    p(\rmH_o \,|\, \gW) = q(\rmH_o),&\quad p(\rmH_t \,|\, \rmH_o, \gW) = \frac{1}{\mathcal{H}} g(\rmH_o,\, \rmH_t)\, q(\rmH_t);\qquad
    \mathcal{H} = \mathbb{E}_{q(\rmH_t)}\left[g(\rmH_o,\,\rmH_t) \right].
    \label{equ: ss-vgae-prior-h}
\end{split}
\end{equation}
It is straightforward to show that the first term in the definition of $\mathcal{L}_{\text{BGCL}}^{w}$ (\eqref{equ: ss-vgae}) is equal to infinite limit of $\mathcal{I}_{\text{NCE}}$ or equivalently $-\mathcal{L}_{\text{CNT}}$ in \eqref{equ: bgrl_loss}.

Finally, we parameterize prior and posterior of $\gW_o$ and $\gW_t$ such that $\KL\left(q(\gW_o) \,||\, p(\gW_o)\right) + \KL\left(q(\gW_t) \,||\, p(\gW_t)\right)$ is proportional to $\mathcal{L}_{\text{WD}}$ in \eqref{equ: bgrl_loss}.
More specifically, we parameterize the posterior of weights as follows:
\begin{equation}
\begin{split}
    q(\gW) = q(\gW_o)\, q(&\gW_t) = \prod_{l=1}^{L} \prod_{(u,v) \in \mathcal{E}} \big[q(\tilde{\rmW}_o^{(l,u,v)})\,\, q(\tilde{\rmW}_t^{(l,u,v)})\big];\\
    q(\tilde{\rmW}_o^{(l,u,v)}) =  \pi_o^{(l)}\, &\delta\left(\tilde{\rmW}_o^{(l,u,v)} - \mathbf{0} \right) + (1 - \pi_o^{(l)})\, \delta\left(\tilde{\rmW}_o^{(l,u,v)}- \mM^{(l)}\right),
    \label{equ: posterior_w}
\end{split}
\end{equation}
where $\delta(\cdot)$ is the Dirac delta function, and $\{\pi_o^{(l)}, \pi_t^{(l)}, \mM^{(l)}\}_{l=1}^L$ are the parameters of the posterior. Note that we use the same form of parametrization for both views and for brevity we wrote the posterior for one view only. We share $\theta_w = \{\mM^{(l)}\}_{l=1}^L$ between two views while having separate drop probabilities for each view.  
To be able to evaluate the KL term analytically, the discrete quantised Gaussian can be adopted as the prior distribution as in \cite{gal2017concrete}.
Then, we can derive the KL divergence between the prior and posterior as:
\begin{equation}
    \KL\left(q(\tilde{\rmW}_o^{(l,u,v)}) \,||\, p(\tilde{\rmW}_o^{(l,u,v)})\right) \propto \frac{(1-\pi_o^{(l)})}{2} \,||\, \mM^{(l)} ||^2 \,-\, \gH(\pi_o^{(l)}),
    \label{equ: kl_bayesian_dl}
\end{equation}
where $\gH$ is the Shannon entropy of Bernoulli random variable. When the augmentation probabilities are fixed, $\gH(\pi_t^{(l)})$ and $\gH(\pi_o^{(l)})$ are constants, hence the KL divergence is only proportional to $\ell^2$ norm of the weights. Therefore, with appropriate choice of $\ell^2$ regularization hyperparameter, it will be equal to $\mathcal{L}_{\text{WD}}$ in \eqref{equ: bgrl_loss}. Our analysis in this subsection showed that training a GCL model with generalized augmentation is an approximation of a VGAE with stochastic encoders. 

The proposed BGCL offers a novel and interesting perspective of contrastive learning. Specifically, it could be interpreted that BGCL has no augmentation over the input and the embeddings are generated by feeding the original data to two different stochastic functions (i.e. moving stochasticity from data to function). Indeed, our unique construction of stochastic encoders allows the model to mimic data augmentation. Whether any other type of stochastic function, such as neural processes, works the same way or not is an interesting avenue for future works.

\subsection{Learnable Bayesian graph contrastive learning}
Finding optimal parameters for augmentation distributions is subject to computationally expensive and time consuming search over a set of possible hyperparameters \cite{you2021graph}. 
Here, we propose a Bayesian approach with variational inference to learn the parameters of our generalized augmentation. 
We impose independent hierarchical beta-Bernoulli priors over augmentations and assume
\begin{equation}
    Z_{o,[i,j]}^{(l,u,v)} \,|\, \pi_{o}^{(l)} \sim  \mathrm{Bernoulli}(\pi_{o}^{(l)}); \quad
    \pi_{o}^{(l)} \sim \mathrm{Beta}(c/L, \, c (L - 1)/L),
    \label{equ: beta_bernoulli}
\end{equation}
where $c$ is a hyperparameter. Furthermore, we parametrize the posterior using Kumaraswamy-Bernoulli hierarchical distribution \cite{kumaraswamy1980generalized} as follows:
\begin{gather}
    q\left(Z_{o,[i,j]}^{(l,u,v)} \,|\, \pi_{o}^{(l)}\right) =  \mathrm{Bernoulli}(\pi^{(l)}); \quad
    q(\pi_{o}^{(l)}) = a_{o}^{(l)}\, b_{o}^{(l)}\, (\pi_{o}^{(l)})^{a_{o}^{(l)} -1}(1 - (\pi_{o}^{(l)})^{a_{o}^{(l)}})^{b_{o}^{(l)} - 1},
    \label{equ: kuma_bernoulli}
\end{gather}
where $a_{o}^{(l)}$ and $b_{o}^{(l)}$ are variational parameters. Note that we use the same form of parameterization for prior and posterior of both views and for brevity we wrote the prior and posterior for one view only. We denote the parameters of our augmentation model by $\theta_a = \theta_{a,o} \cup \theta_{a,t} = \{a_o^{(l)}, b_o^{(l)}\}_{l=1}^L \cup \{a_t^{(l)}, b_t^{(l)}\}_{l=1}^L$. Furthermore, we represent the set of latent augmentation masks by $\mathcal{Z} = \mathcal{Z}_o \cup \mathcal{Z}_t = \{\{\rmZ_o^{(l,u,v)}\}_{(u,v) \in \gE}\}_{l=1}^L \cup \{\{\rmZ_t^{(l,u,v)}\}_{(u,v) \in \gE}\}_{l=1}^L$, and the set of augmentation drop rates by $\boldsymbol{\Pi} = \boldsymbol{\Pi}_t \cup \boldsymbol{\Pi}_o = \{\pi_t^{(l)}\}_{l=1}^L \cup \{\pi_o^{(l)}\}_{l=1}^L$.

The main goal here is learning ``good'' augmentations which should be reflected in the likelihood function. However, before defining the likelihood, we have to define what makes ``good'' augmentations. \citet{tian2020makes} argues that a good set of views are those that share the minimal information necessary to perform well at the downstream task which is equivalent to \emph{maximizing} contrastive loss.
Following the same principle, we define a likelihood that places more mass on views with lower mutual information. 
Let's consider the same VGAE that was introduced in the last subsection. While there we assumed that weights are random variable with fixed augmentation drop rates, here we assume that $\theta_w = \{\mM^{(l)}\}_{l=1}^{L}$ are fixed and $\theta_a$ are variational parameters. Following the same analysis as last section, we can write the ELBO for this model as follows:
\begin{equation}
\begin{split}
    \mathcal{L}_{\text{BGCL}}^{a} &= \mathbb{E}_{(\rmH_o, \rmH_t, \gZ, \boldsymbol{\Pi} \,|\, \gG)} \log \left[\frac{p(\rmH_o, \rmH_t \,|\, \gZ, \boldsymbol{\Pi})}{q(\rmH_o)\, q(\rmH_t) } \right]\\
    &\quad - \KL\left(q(\gZ_o,\,\boldsymbol{\Pi}_o) \,||\, p(\gZ_o,\,\boldsymbol{\Pi}_o)\right) - \KL\left(q(\gZ_t,\,\boldsymbol{\Pi}_t) \,||\, p(\gZ_t,\,\boldsymbol{\Pi}_t)\right) + \log\, p_{\text{true}}(\gG).
    \label{equ: ss-vgae-a}
\end{split}
\end{equation}

Let's assume that the prior of latent representation are defined as follows:
\begin{equation}
\begin{split}
    p(\rmH_o \,|\, \gZ, \boldsymbol{\Pi}) = q(\rmH_o),&\qquad p(\rmH_t \,|\, \rmH_o, \gZ, \boldsymbol{\Pi}) = \frac{1}{\mathcal{H}}  g(\rmH_o,\, \rmH_t)^{-1}\, q(\rmH_t);\\
    &\mathcal{H} = \mathbb{E}_{q(\rmH_t)}\left[g(\rmH_o,\,\rmH_t)^{-1} \right].
    \label{equ: ss-vgae-prior-h-a}
\end{split}
\end{equation}
By substituting the above priors in \eqref{equ: ss-vgae-a}, and using Jensen's inequality, we can see that the first term in $\mathcal{L}_{\text{BGCL}}^{a}$ is bounded by the infinite limit of $-\mathcal{I}_{\text{NCE}}$ or equivalently $\mathcal{L}_{\text{CNT}}$.
Furthermore, we can expand the KL terms in $\mathcal{L}_{\text{BGCL}}^{a}$ as follows:
\begin{equation*}
\begin{split}
&\KL\left(q(\gZ_o,\,\boldsymbol{\Pi}_o) \,||\, p(\gZ_o,\,\boldsymbol{\Pi}_o)\right)
= \sum_{l=1}^{L_o} \sum_{(u,v)\in\mathcal{E}} \KL\left(q(\mathbf{Z}_o^{(l,u,v)}, \pi_o^{(l)}) \,||\, p(\mathbf{Z}_o^{(l,u,v)}, \pi_o^{(l)})\right) \\
&\,\,= \sum_{l=1}^{L_o} \sum_{(u,v) \in \mathcal{E}} \KL\left(q(\mZ_o^{(l,u,v)} \,|\, \pi_o^{(l)}) \,||\, p(\mZ_o^{(l,u,v)} \,|\, \pi_o^{(l)})\right) + \KL\left(q(\pi_o^{(l)}) \,||\, p(\pi_o^{(l)})\right).
\end{split}
\end{equation*}
Since the conditional posterior and prior of masks have the same distribution, i.e. Bernoulli, the $\mathrm{KL}$ divergence between them is zero. Hence, the $\mathrm{KL}$ term only depends on the $\mathrm{KL}$ divergence between posterior and prior of augmentation drop rates, which can be expanded as follows:
\begin{equation*}
\begin{split}
     \KL&\left(q(\pi_o^{(l)}) \,||\, p(\pi_o^{(l)})\right) =
     \sum_{l=1}^{L} \Big[\frac{a_o^{(l)} - c_o/L}{a_o^{(l)}}\left(- \gamma - \Psi(b_o^{(l)}) - \frac{1}{b_o^{(l)}} \right)
     + \mathrm{log} \frac{a_o^{(l)} b_o^{(l)}}{c_o/L} - \frac{b_o^{(l)} - 1}{b_o^{(l)}}\Big],
\end{split}
\end{equation*}
where $c_o$ is a hyperparameter, $\gamma$ is the Euler-Mascheroni constant and $\Psi(\cdot)$ is the digamma function. Note that for brevity we wrote the last two equations for one view only.

Due to the discrete nature of the random masks $\gZ$, we cannot directly apply reparameterization trick to calculate the gradient of the $\mathrm{KL}$ term. Hence, we use continuous approximation, i.e. concrete relaxation \cite{jang2016categorical,gal2017concrete}, of $\gZ$ instead of discrete variables. 
Given a drop probability $\pi$ and a sample from uniform distribution $u \sim \mathrm{Unif}[0, 1]$, a sample from concrete distribution can be calculated as follows:
\begin{equation*}
    \mathrm{sigmoid}\left(\frac{1}{t}\left( \mathrm{log}\big(\frac{\pi}{1 - \pi}\big) + \mathrm{log}\big(\frac{u}{1-u}\big)\right)\right),
\end{equation*}
where $t$ is temperature parameter. We emphasize that to the best of our knowledge, this is the first method that models augmentations as discrete random variables which offers modeling more diverse types of augmentation.

Having defined the augmentation learning method, we describe the overall learning procedure. We have two objective functions: 1) the objective function for contrastive model (\eqref{equ: bgrl_loss} or equivalently \eqref{equ: ss-vgae}) which is used for learning the parameters of encoders $\theta_w$, and 2) the ELBO for our Bayesian augmentation learning method (\eqref{equ: ss-vgae-a}) which is used for learning the parameters of augmentation $\theta_a$. Combining these objective functions, we derive the overall objective of our model as:
\begin{equation}
\begin{split}
    \min_{\theta_w}&\quad \mathcal{L}_{\text{CNT}} + \mathcal{L}_{\text{WD}}\\
    \max_{\theta_a}&\quad \mathcal{L}_{\text{CNT}} - \KL\left(q(\gZ,\,\boldsymbol{\Pi}) \,||\, p(\gZ,\,\boldsymbol{\Pi})\right)\,
\end{split}
\end{equation}
In each training epoch, we first optimize the parameters of encoders by minimizing the contrastive loss. Then we optimize the parameters of augmentations by maximizing the ELBO of our Bayesian augmentation learning method.

\subsection{Downstream predictive model}

In \secref{sec: elbo}, by performing varitional inference for BGCL, we 
infer the posterior distribution of weights of model.
By sampling from posterior of the weights of the model and passing the input graph to the model, we will have samples from the approximate posterior of node representations. In fact, these are samples from an implicit variational approximation $Q(\rmH)$ of the true posterior distribution. 
Let's consider the node classification task where $\rvy = \{y_v\}_{v \in \mathcal{V}}$ are the set of observed node labels in the graph. We model the following probability distribution:
\begin{equation}
    \rvy \sim p_{\mathrm{cat}}(\rvy;\, \mathrm{softmax}(\rmH\,\mW_{c})); \qquad  \rmH \sim Q(\rmH) \nonumber
\end{equation}
where $\rmH$ is the matrix of node representations, $\mW_c$ is the parameters of the multinomial logistic regression, and $p_{\mathrm{cat}}$ is probability mass function of categorical distribution.
We can calculate the likelihood of $\rvy$ as follows:
\begin{equation}\label{bayes_likelihood}
  p(\rvy) = \int p_{\mathrm{cat}}(\rvy;\, \mathrm{softmax}(\mH\,\mW_{c}))\, Q(\rmH) \,d\mH.
\end{equation}
Similar formulation can also be obtained for the regression task by replacing the categorical distribution with normal density with a fixed variance and ignoring softmax function for the mean.

In the non-Bayesian contrastive learning setup, a common practice for learning the parameters $\mW_c$ is via maximum log-likelihood. We also take the same approach, however, we do not have a closed-form likelihood. 
We circumvent this problem by using the Monte Carlo estimate of the likelihood (equation \ref{bayes_likelihood}) as:
\begin{equation}\label{bayes_likelihood_monte}
 \frac{1}{K} \sum_{i=1}^{K} p_{\mathrm{cat}}(\rvy;\, \mathrm{softmax}(\mH^{(i)}\,\mW_{c})); \quad \quad \mH^{(i)}  \:  \stackrel{iid}{\sim} \: Q(\rmH) 
\end{equation}
and learn parameters of logistic regression $\mW_c$ by maximizing the logarithm of the above term via batch-optimization. 
Our model is more flexible with higher expressive power compared to the existing methods. 
More specifically, based on equation \ref{bayes_likelihood_monte}, the proposed framework learns a mixture of categorical (normal) distributions rather than a single one for classification (regression) task. Therefore, it is less prone to misspecification, providing more reliable predictions.

\section{Experiments}
\label{sec:experiments}

\paragraph{Datasets.}
We use three standard graph datasets to benchmark the performance of BGCL as well as baselines. Cora \cite{sen2008collective}, Citeseer \cite{sen2008collective}, and Amazon-Photos \cite{mcauley2015image}. 
Cora and Citeseer are citation datasets where each node represents a document and the edges indicate the citation relations. 
Each node is provided with a multi-dimensional binary attribute vector, where each dimension of the attribute vector indicates the presence of a word from a dictionary in the publication. 
We follow the same pre-processing steps as \cite{zhu2020grace} and randomly split the nodes into train (10\%) and test (90\%) sets.
Amazon-Photo is from the Amazon co-purchase graph where nodes represent products and edges are between pairs of products frequently purchased together. Products are classified into eight classes, and node attributes are a bag-of-words representation of a product’s reviews. 
We follow the same pre-processing steps as \cite{thakoor2021bootstrapped} and randomly split the nodes into train/validation/test sets (10/10/80\%), respectively.

\paragraph{Baselines and experimental setups.}
Since GRACE achieves the state-of-the-art results in graph contrastive learning \cite{thakoor2021bootstrapped}, we use it as the base contrastive model in our BGCL. More specifically, we use our learnable Bayesian generalized augmentation approach on top of GRACE. We emphasize that our proposed model could also be used on top of any graph contrastive model which deploys $\sT$ as the set of augmentation types. For a fair comparison with baselines, we use the same set of hyperparameters as the ones reported in GRACE original publication \cite{zhu2020grace} for BGCL except for the augmentation drop rates which are learned in the training process.
The prior distribution for all of the drop rates, $\pi^{(l)}$s, is $\mathrm{Beta}(1, 1)$. The temperature in concrete distribution is 0.3.
The linear classifiers are trained for 50 different random splits and random initializations. We implemented our model in PyTorch \cite{pytorch}. All of our simulations are conducted on a single NVIDIA Tesla P100 GPU node.

We compare our BGCL with several baselines 
in terms of node classification accuracy. More specifically, we compare our BGCL with two contrastive methods (\textbf{GRACE} \cite{zhu2020grace} and \textbf{DGI} \cite{sun2019infograph}), two auto-encoders (\textbf{VGAE} and \textbf{GAE} \cite{kipf2016variational}), two conventional node embedding methods (\textbf{node2vec} \cite{grover2016node2vec} and \textbf{DeepWalk} \cite{perozzi2014deepwalk}), and \textbf{supervised GCN} \cite{kipf2016semi}. We use the same setup for baselines as described in \cite{zhu2020grace}. Since there are no contrastive models with uncertainty quantification capabilities, we compare our BGCL with a supervised model, Bayesian GCN with \textbf{Graph DropConnect (GDC)} \cite{hasanzadeh2020gdc}. 
We use the same setup as \cite{hasanzadeh2020gdc} to quantify uncertainty for datasets. More details on experimental setups and datasets are included in the supplementary materials.

\begin{table}[t]
	\centering
	\caption{Performance of our BGCL and baselines on node classification in terms of accuracy (in \%). The highest performance of unsupervised models is highlighted in boldface.}
	\label{tab: class_acc}
	\vspace{4pt}
    \resizebox{0.65\columnwidth}{!}{%
	\begin{tabular}{ccccc}
	\toprule
	Method & Cora & Citeseer & Amazon-Photo \\
	\midrule
	Raw features & 64.8 & 64.6 & 78.53 \\
	node2vec & 74.8 & 52.3 & 89.72 \\
	DeepWalk & 75.7 & 50.5 & 89.44 \\
	DeepWalk + features & 73.1 & 47.6 & 90.05 \\
	\midrule
	GAE    & 76.9 & 60.6 & 91.68 \\
	VGAE   & 78.9 & 61.2 & 92.24 \\
	DGI    & 82.6{\footnotesize \textpm0.4} & 68.8{\footnotesize \textpm0.7} & 91.61{\footnotesize \textpm0.22} \\
	GRACE  & 83.3{\footnotesize \textpm0.4} & 72.1{\footnotesize \textpm0.5} & 92.15{\footnotesize \textpm0.24} \\
	\textbf{BGCL}  & \textbf{83.77{\footnotesize \textpm0.28}} & \textbf{72.71{\footnotesize \textpm0.25}} & \textbf{92.51{\footnotesize \textpm0.15}} \\
	\specialrule{0.5pt}{0.5pt}{1pt}
	\midrule
	Supervised GCN  & 82.8 & 72.0 & 92.42 \\
	\bottomrule
	\end{tabular}
	}
\end{table}

\subsection{Uncertainty quantification in node classification}
Ideally, if a model is confident about its prediction, the prediction
should be accurate (i.e. high $p(\text{accurate}\,|\,\text{certain})$). Also, if
a model makes a mistake in the prediction, it should be uncertain
about its output (i.e. high $p(\text{uncertain}\,|\,\text{inaccurate})$). These two probabilities can be combined into one to form metric for quality of uncertainty called Patch Accuracy Vs Patch Uncertainty (PAVPU) \cite{mukhoti2018evaluating}. More specifically, $\text{PAVPU} = (n_{ac}+n_{iu})/(n_{ac}+n_{au}+n_{ic}+n_{iu})$, where $n_{ac}$ is the number of accurate and certain predictions, $n_{au}$ is the number of accurate and uncertain predictions, $n_{ic}$ is the number of inaccurate and certain predictions, and $n_{iu}$ is the number of inaccurate and uncertain predictions. A model with higher PAVPU quantifies uncertainty more accurately. 
We use predictive entropy (as defined in \cite{mukhoti2018evaluating}) as a measure for certainty. 
\begin{wrapfigure}{r}{0.5\textwidth}
  \begin{center}
    \includegraphics[width=0.49\textwidth]{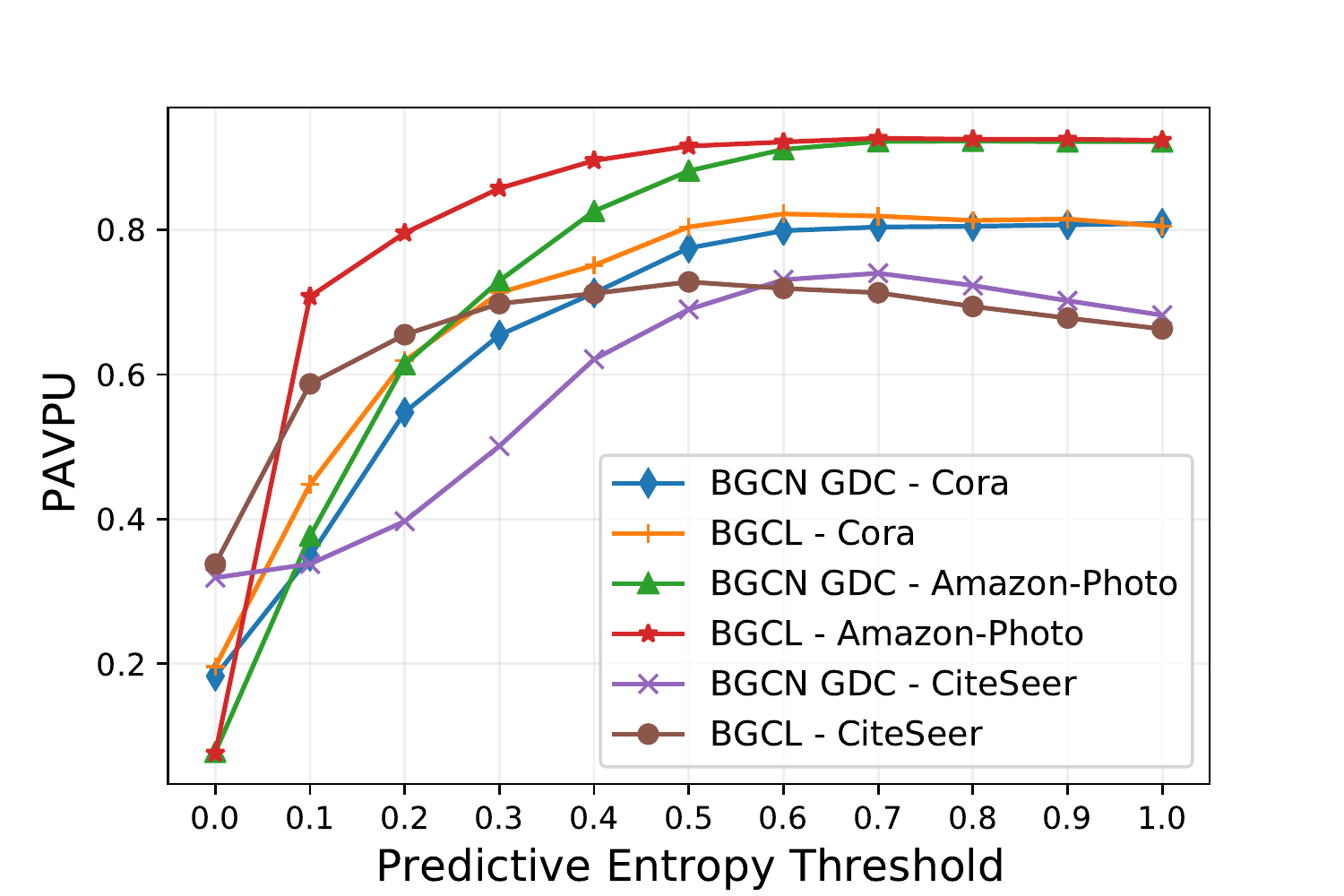}
  \end{center}
  \caption{Comparison of uncertainty estimates in terms of PAvPU by a 128-dimensional 4-layer Byesian GCN with GDC, 
    and BGCL on Cora, Citeseer and Amazon-Photo. 
    }%
    \label{fig: uncertain}
    \vspace{-0.6cm}
\end{wrapfigure}

Figure \ref{fig: uncertain} shows performance of BGCL and a supervised Bayesian GCN model in uncertainty quantification on benchmark datasets. Comparing these two models, we see that BGCL is outperforming (or performs on par with) competing method 
on most of the certainty thresholds. This is indeed very impressive considering that BGCL is a \emph{self-supervised} model and baseline is a \emph{supervised} model. This empirically proves that our novel Bayesian formulation of graph contrastive learning allows us to accurately capture uncertainty of latent representations.

PAVPU alone may not depict the full picture of performance of a model. Therefore, we also report the classification accuracy for our model and compare it with the existing baselines in Table \ref{tab: class_acc}. The results show that BGCL outperforms the unsupervised and self-supervised baselines in all datasets.
We emphasize that our goal for reporting classification accuracy is not claiming state-of-the-art performance in node classification. Rather, we want to show that our learnable generalized augmentation scheme is able to learn optimal drop rates without any supervision and obviates the need for ad-hoc procedure for selecting augmentation type. Indeed there are other \emph{supervised} methods that outperform contrastive models. However, the main focus of this paper is self-supervised setting and the reported supervised GCN (like majority of works in graph contrastive learning) is merely for showing the relative performance of self-supervised techniques with a well-known supervised model.

\subsection{Representation uncertainty}\label{app: rep_uncertainty}
We provide results showing how uncertainty of representations increases the interpretability of our model. Specifically, we show that adding noise to input data (which can be seen as out of distribution samples), will increase the uncertainty of representation.

\begin{figure}
     \centering
     \begin{subfigure}[b]{0.48\textwidth}
         \centering
         \includegraphics[width=\textwidth]{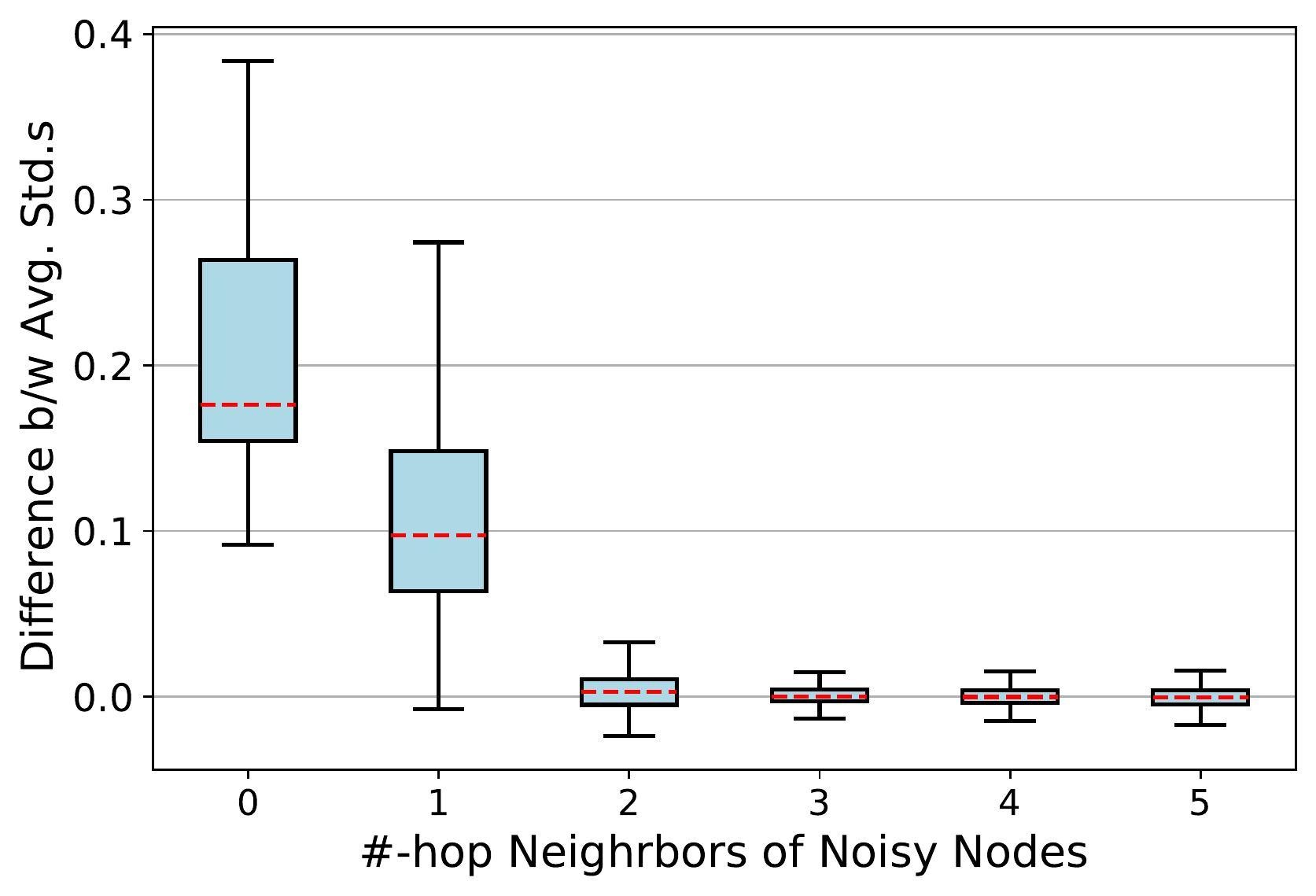}
         \caption{Cora}
         \label{fig: cora_std}
     \end{subfigure}
     \hfill
     \begin{subfigure}[b]{0.48\textwidth}
         \centering
         \includegraphics[width=\textwidth]{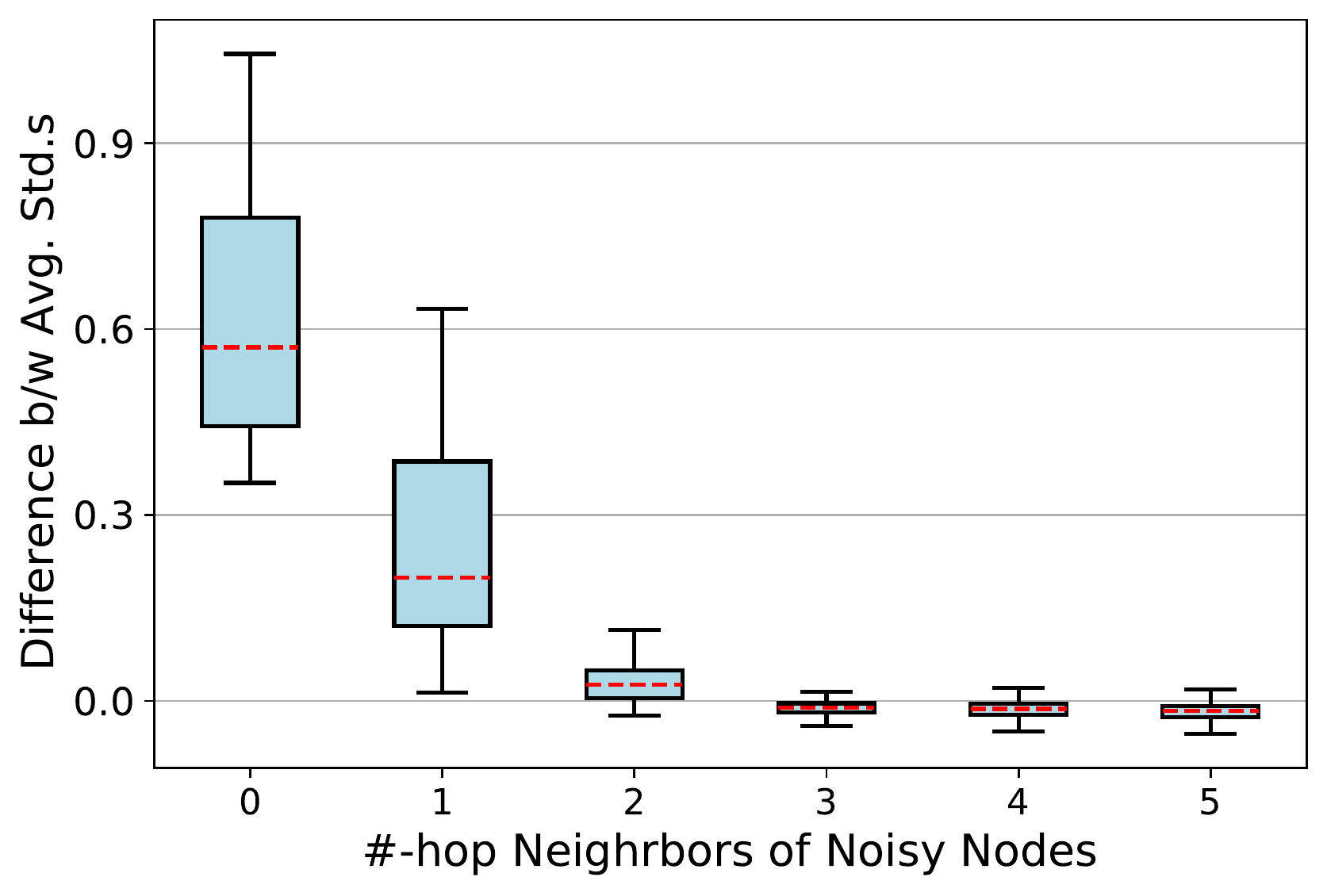}
         \caption{Citeseer}
         \label{fig: citeseer_std}
     \end{subfigure}
    \caption{Effect of out-of-distribution (o.o.d) node attributes on representation uncertainty.}
\end{figure}

To that end, we first train three BGCL models on Cora, Citeseer and Amazon-photo datasets. Then, we synthesize noisy versions of these datasets by randomly selecting a few nodes (10 for Cora and Citeseer, and 100 for Amazon-photo), and replace their attributes with normally distributed noise ($\mathcal{N}(0,1)$ for Cora and Citeseer, and $\mathcal{N}(0,1)$ and $\mathcal{N}(0,10)$ for Amazon-photo). We further train three other BGCL models on these noisy datasets. 
Then, we examine the difference between the Average Standard Deviation (ASTD) of latent representations of the randomly selected nodes in the two cases for each dataset. We do the same examination for $k$-hop neighbors (for $k=1,\cdots,5$) of the randomly selected nodes as well. 
Note that given $s$ Monte-Carlo samples from $d$-dimensional latent representation of a node (denoted by a $d \times s$ matrix), the ASTD is calculated by first calculating the standard deviation along the second dimension and then averaging over the first dimension. 

\begin{wrapfigure}{r}{0.5\textwidth}
    \centering
    \includegraphics[width=0.49\textwidth,keepaspectratio]{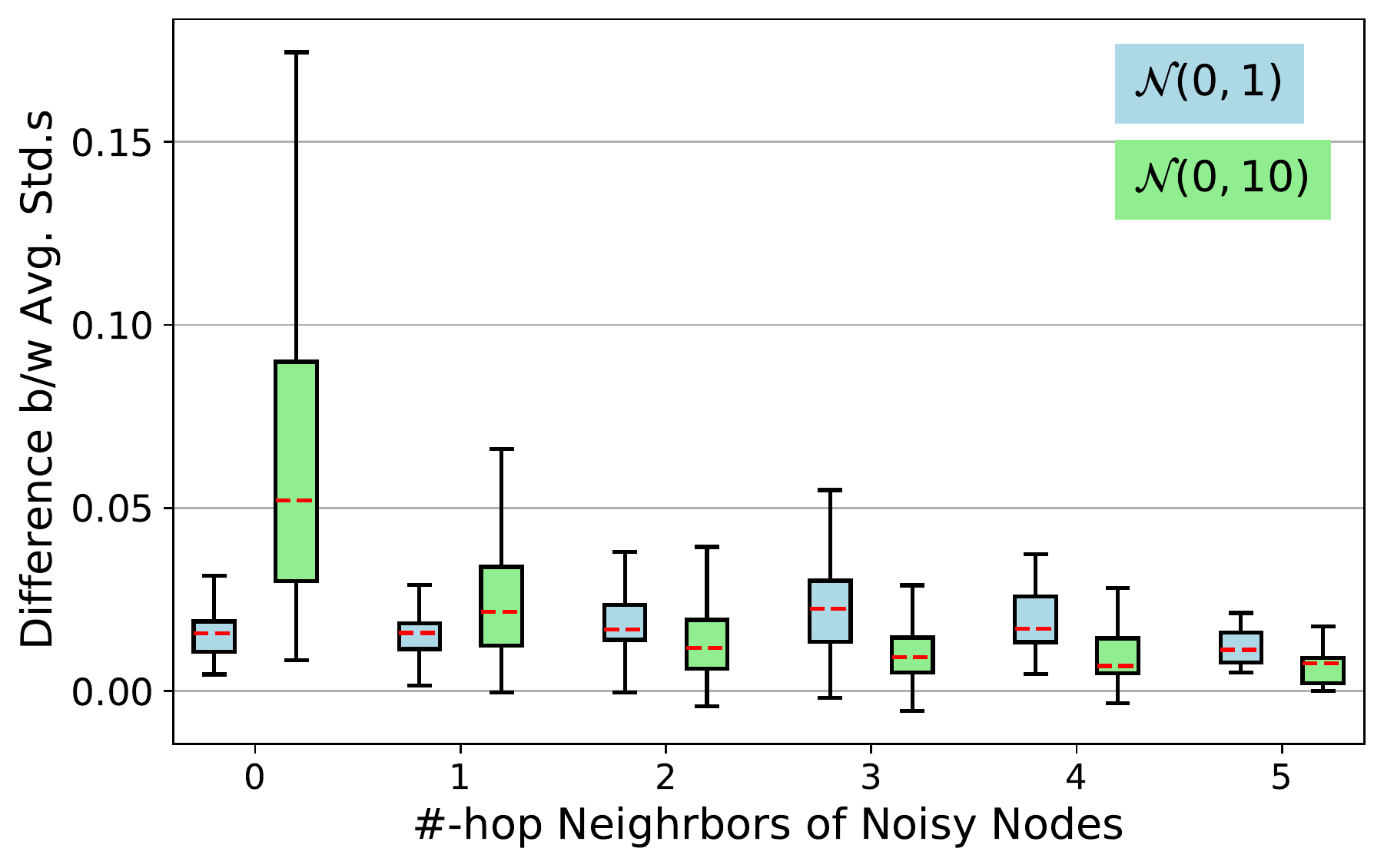}
    \caption{Effect of o.o.d node attributes on representation uncertainty for Amazon-Photo.}
    \label{fig: amazon_std}
\end{wrapfigure}

Figures \ref{fig: cora_std} and \ref{fig: citeseer_std} show the box plot of the difference between ASTDs of latent representations of randomly selected nodes (as well as their $k$-hop neighbors) in noisy and noise-free versions of Cora and Citeseer. Note that 0-hop neighbors refers to the randomly selected nodes themselves. We can see that the noisy nodes have the highest increase in ASTD. 
The reason for higher ASTD in noisy cases is that information mismatch between the attribute of noisy nodes and their neighbors.
As we go farther away from the noisy nodes, the difference gets closer to zero. 
Noting that GCNs \cite{kipf2016semi} are low-pass graph filters, this observation is consistent with evolution of a diffusion process over graph. 
Figure \ref{fig: amazon_std} shows the results for Amazon-photo dataset. We can see that adding $\mathcal{N}(0,1)$ does not effect the ASTD significantly. This can be explained by higher average degree of Amazon-photo (31.13) compared to Cora (3.90) and Citeseer (2.77). Having more neighbors allows the model to filter out noise, by taking advantage of the noise-free attributes of the neighbors via message passing, and learns embeddings with higher confidence. By increasing the standard deviation of noise to 10, we reduce the signal-to-noise ration showing an increase in uncertainty of the latent representations. 
This is consistent with our observation for Cora and Citeseer.

\section{Conclusions}
In this paper, we proposed Bayesian Graph Contrastive Learning (BGCL). First, we introduced a generalized augmentation method by applying random augmentation to each layer of encoders, obviating the need to choose augmentation type. Next, we show that a graph contrastive learning model with the proposed augmentation method, is an approximation of a VGAE with Bayesian neural networks as encoders. As a result, our BGCL represents each node by a distribution in the latent space instead of deterministic node embeddings. Hence, BGCL provides predictive uncertainty in downstream graph analytics tasks. 
Furthermore, we developed a Bayesian framework to infer the augmentation drop rates in each view of the contrastive model, eliminating the need for a computationally expensive search for hyperparameter tuning.

\bibliography{graphbib.bib}

\newpage

\appendix

\section{Further results on learnable augmentation}\label{app: more_results}
While the main focus of the experiments/discussions in the main manuscript was uncertainty quantification, here, we show that our learnable generalized augmentation can be deployed in non-Bayesian scenarios as well. 
To deploy our proposed model in a non-Bayesian setting, we train our model as before. However, in testing phase, we don't sample from generalized augmentation. Instead, we feed the input graph to encoder without augmentation. This yields \emph{deterministic} node representations. These deterministic representations could be used in downstream tasks in the same way as other graph contrastive models. 

We report the results of our model in non-Bayesian setting for three additional datasets: WikiCS \cite{pennington2014glove}, Amazon-Computers \cite{mcauley2015image}, and CoauthorCS \cite{sinha2015an}. The details of datasets are reported in Table \ref{tab: data_stats}. We use cluster batching \cite{chiang2019clustergcn} (as implemented in PyTorch Geometric package) during training of our model on these datasets. The number of batches for WikiCS, Amazon-Computers, and CoauthorCS are 5, 5, and 10, respectively. Rest of hyperparameters are reported in Table \ref{tab: hyperparameters}. For these datasets, we use the same experimental settings as \cite{thakoor2021bootstrapped}, and report the baseline results from \cite{thakoor2021bootstrapped}.

\begin{table}[!h]
\centering
\caption{Performance of BGRL and baselines in non-Bayesian Setting.}
\vspace{0.05in}
\begin{tabular}{lccccc}
\hline
& WikiCS   & Amazon-Computers & CoauthorCS\\
\hline                                                          
Raw features & 71.98 $\pm$ 0.00 & 73.81 $\pm$ 0.00 &  90.37 $\pm$ 0.00 \\
DeepWalk & 74.35 $\pm$ 0.06 & 85.68 $\pm$ 0.06 &  84.61 $\pm$ 0.22 \\
DeepWalk + feat. & 77.21 $\pm$ 0.03 & 86.28 $\pm$ 0.07 & 87.70 $\pm$ 0.04 \\
\hline
DGI  & 75.35 $\pm$ 0.14 & 83.95 $\pm$ 0.47  & 92.15 $\pm$ 0.63 \\
GRACE  & 80.14 $\pm$ 0.48 & 89.53 $\pm$ 0.35  & 91.12 $\pm$ 0.20 \\
BGCL (non-Bayesian) & 80.62 $\pm$ 0.35 & 89.70 $\pm$ 0.36 & 91.72 $\pm$ 0.23\\
\hline
\bottomrule
\end{tabular}
\label{tab: results_non_bayes}
\end{table}

The results, reported in Table \ref{tab: results_non_bayes}, show that our learnable generalized augmentation achieves on-par performance with the base model (GRACE) without any need for an expensive grid search to find optimal augmentation type or augmentation drop rates in a non-Bayesian setting.

\section{Sampling complexity}\label{app: limitations}
As described in the main manuscript, in our generalized augmentation model, we need to draw $F_l \times F_{l+1} \times E$ with $E$ as number of edges. This could potentially be a large number. To alleviate this issue, we use a single sample for a block of features as opposed to drawing a new sample for every input-output feature pairs. This reduces the number of samples to $B \times E$ with $B$ being number of blocks. 
The number of blocks used for each experiment is reported in Table \ref{tab: hyperparameters}.

\section{Further implementation details}\label{app: imp_details}
All of the datasets used in our experiments are publicly available in PyTorch Geometric \cite{fey2019torchgeometric} package. Table \ref{tab: hyperparameters} shows the hyperparameters used in our experiments for each dataset. $\mathrm{LR}_w$ in Table \ref{tab: hyperparameters} refers to learning rate for the first step of optimization while $\mathrm{LR}_a$ denotes the learning rate for the second step of optimization. Note that, for a fair comparison, hidden dimension, latent dimenstion, activation function, and $\tau$ are the same as experiments in GRACE \cite{zhu2020grace}. For simplicity, we assumed that the augmentation in all layers of a view have the same drop rate. All models are initialized with Xavier initialization \cite{glorot2010init}, and trained using Adam optimizer \cite{kingma2014adam}. For down-stream classification task, we have implemented our linear classifier in PyTorch. For all of the datasets, we train the classifier for 150 epochs using Adam optimizer with the learning rate of 0.1.

To calculate PAVPU, we first draw 500 random augmentations from the trained BGCL model leading to 500 embeddings for each node.
Next, we train a classifier (as described in section 3.3) using 10 samples. Then, we deploy the trained classifier to predict the nodes' labels using the remaining 49 sets of 10 samples. Having 49 prediction for each node, we follow the same procedure as in \cite{mukhoti2018evaluating} to calculate PAVPU.

\begin{table}[!h]
\vspace{-0.01in}
\centering
\caption{Graph dataset statistics.}
\vspace{0.05in}
\begin{tabular}{@{}l | c c c c c}
\toprule
Dataset & Nodes & Edges & Features & Classes\\ \midrule
\textbf{Cora} & 2,708 & 5,429 & 1,433 & 7\\
\textbf{Citeseer}  & 3,327 & 4,732 & 3,703 & 6 \\
\textbf{Amazon-Photo} & 7,650 & 119,081 & 745 & 8 \\
\textbf{Amazon-Computers} & 13,752 & 245,861 & 767 & 10 \\
\textbf{WikiCS} & 11,701 & 216,123 & 300 & 10 \\
\textbf{CoauthorCS} & 18,333 & 81,894  & 6,805 & 15 \\
\bottomrule
\end{tabular}
\label{tab: data_stats}
\end{table}

\begin{table*}[!h]
	\small
	\centering
	\caption{Hypeparameters used in the experiments.}
	\resizebox{0.95\columnwidth}{!}{%
    \begin{tabular}{cccccccccc}
	\toprule
	Dataset & $\mathrm{LR}_w$ & $\mathrm{LR}_a$ & $\ell_2$ &  \# of blocks & Epochs & Hidden dim & Latent dim & Activation & $\tau$ \\
	\midrule
	Cora  & 0.0005 & 0.001 & \(5 \times 10^{-9}\) & 8 & 250   & 256 & 128 & ReLU & 0.4\\
	Citeseer & 0.001 & 0.0005 & \(5 \times 10^{-9}\) & 8 & 250   & 512 & 256   & PReLU & 0.9\\
	Amazon-Photo & 0.001 & 0.0005 & \(5 \times 10^{-9}\) & 2 & 250  & 512 & 256   & PReLU & 0.7\\
	Amazon-Computers & 0.0001 & 0.0005 & \(5 \times 10^{-9}\) & 32 & 50  & 512 & 256   & ReLU & 0.4\\
	WikiCS & 0.0001 & 0.0005 & \(5 \times 10^{-9}\) & 16 & 50  & 512 & 256   & ReLU & 0.4\\
	CoauthorCS & 0.0001 & 0.0005 & \(5 \times 10^{-9}\) & 16 & 50  & 256 & 128   & ReLU & 0.4\\
	\bottomrule
	\end{tabular}
	}
	\label{tab: hyperparameters}
\end{table*}

\section{Broader impacts}\label{app: broader_impacts}
One often-neglected side effect of improving the interpretability of a machine learning model is the risk of automation bias and a false sense of trust in models. The ease of understanding the system
increases the risk of misinterpretation, overtrust, and even incorrect use of the outputs. Higher accuracy and interpretability of our model may further increase the chance of that risk.

\end{document}